# Semantic Approach to Quantifying the Consistency of Diffusion Model Image Generation

Brinnae Bent, PhD
Duke University
Durham, NC, USA
brinnae.bent@duke.edu

**Abstract**

*In this study, we identify the need for an interpretable, quantitative score of the repeatability, or consistency, of image generation in diffusion models. We propose a semantic approach, using a pairwise mean CLIP (Contrastive Language-Image Pretraining) score as our semantic consistency score. We applied this metric to compare two state-of-the-art open-source image generation diffusion models, Stable Diffusion XL and PixArt-α, and we found statistically significant differences between the semantic consistency scores for the models. Agreement between the Semantic Consistency Score selected model and aggregated human annotations was 94%. We also explored the consistency of SDXL and a LoRA-fine-tuned version of SDXL and found that the fine-tuned model had significantly higher semantic consistency in generated images. The Semantic Consistency Score proposed here offers a measure of image generation alignment, facilitating the evaluation of model architectures for specific tasks and aiding in informed decision-making regarding model selection.*

## 1. Introduction

As the study and application of image generation with diffusion models grows, there is a need for better interpretability, explainability, and understandability of the outputs generated by these models [1].

When applying diffusion models to image generation, there is variability in the generated output. This variability is a result of stochastic elements that are inherent to the process of diffusion, including random initialization of the diffusion process, sampling from probability distributions, and nonlinear activations [2].

While variability is inherent to diffusion models, there are differences in the level of variability across different models, due to differences in model architecture, training procedures (including approximations), and techniques used to control or guide the generation process [3]. When applying these models to real world problems, the challenge lies in reconciling the desire for diversity and creativity in generated outputs with the need for consistency and coherence relative to the input prompt.

Quantifying the consistency, or repeatability, of generated outputs would allow for the quantification of this variability and would enable decisions to be made on this reconciliation between creativity and consistency when deciding on a diffusion model for a particular application or task. This quantification can provide an assessment of model stability and consistency, detect unintended bias, validate interpretations of model outputs, and enhance user understanding.

## 2. Related Work

It is commonplace in image generation research to repeat experiments due to "random variation" during inference in diffusion models [4]. However, the number of times an experiment is repeated is often arbitrary because there is no score to assess the effect of this random variation on repeatability or consistency of image generation tasks. This repeatability is likely model dependent, but this has not been studied.

### 2.1. Measuring Image Quality

There has previously been work to measure similarity between two different images, with the primary use case being to assess quality of generated images, like the Inception Score (IS), the Fréchet Inception Distance (FID), and the Kernel Inception Distance (KID) which evaluate the distribution of generated images or their features, and in FID and KID, compare it with the distribution of a set of real images or their features [5]. While these methods have been used extensively in image quality assessment, they have not been used to assess the consistency of image generation models. Further, these methods may be challenging to use when explaining consistency of image generation models due to their lack of direct relevance to human perception of image quality or repeatability, compared to a semantic approach.

### 2.2. Semantic Evaluations

Recently, research into the evaluation of consistency image-text matching in image generation has explored a semantic approach using embeddings of text and images in multimodal embedding models [6-8]. CLIPScore, which uses CLIP to assess image-caption compatibility without using references, achieved high correlation with human judgements [8]. Semantic approaches have also been used to calculate loss for style-consistent image synthesis [9] and in measuring the success of diffusion models in imitating human artists [4].

## 3. Methods

### 3.1. Semantic Consistency Score

We have identified the need for a score to quantify the repeatability or consistency of image generation in diffusion models. In this paper, we propose a semantic approach to this score, using a pairwise mean CLIP score (Equation 1).

$$SCS = \frac{1}{N(N-1)/2} \sum_{i=1}^{N} \sum_{j=i+1}^{N} \max(100 \times cos(E_i, E_j), 0) \quad (1)$$

Equation 1 shows our Semantic Consistency Score, which is a pairwise mean CLIP score, where N is the number of images, $E_i$ and $E_j$ are the CLIP visual embeddings for images $i$ and $j$, respectively. For better interpretability and understanding, the score is bound between 0 and 100, with scores closer to 100 indicating more semantically consistent generated images. The summation of all pairwise cosine similarities is divided by the total number of unique image pairs. The mean is used to ensure that the metric is sensitive to outliers.

CLIP [10] is a cross-modal retrieval model. It was trained on 400M (image, caption) pairs from 500K web search queries. Similar to the original CLIP Score [8], the CLIP model used to compute the Semantic Consistency Score is the ViT-B/32 version, which uses a Vision Transformer [11, 12]. This network outputs a single vector representing the content of the image with 512 dimensions. The weights of the model are trained to maximize the scaled cosine similarity of true image/caption pairs and minimize the similarity of mismatched image/caption pairs, creating an embedding space that has been used for a wide range of applications, from image captioning [8] to image retrieval [13] and image search [14].

### 3.2. Evaluation of Image Generation Models

We wanted to evaluate the consistency of state-of-the-art image generation models. We decided to compare SDXL [15] and PixArt-α [16] because both the weights and architecture are open source. Closed image generation models including DALL-E 3, Imagen 2, and Midjourney are challenging to use for a consistency study because of a lack of transparency around image generation and the inability to set a random seed, which we believe is important for study repeatability. Another issue that further motivated our decision against using DALL-E 3 for this analysis was that the API rewrites prompts, so neither prompts nor random seeds could be set to remove confounding variables when running our experiments.

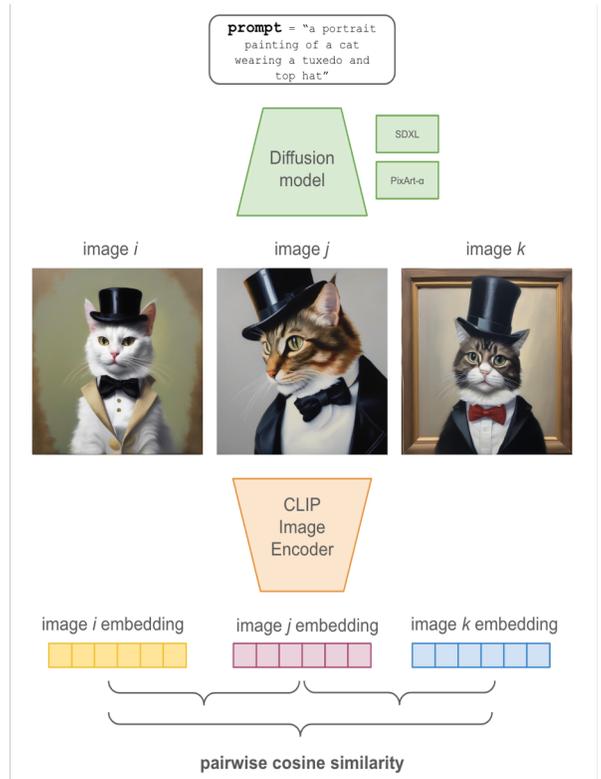

Figure 1: A single prompt is passed through a diffusion model $n$ times with pre-set random seeds. Generated images are passed through the CLIP Image Encoder and a pairwise cosine similarity is taken between the embeddings of all images generated from a single prompt.

Figure 1 shows our proposed approach to evaluate large image generation models. First, a prompt is passed to an image generating diffusion model (SDXL or PixArt-α). This is repeated for $n$ images. Image embeddings are created using the CLIP image encoder. Finally, the pairwise cosine similarity is computed and averaged to be the final score for the given prompt and model.

### 3.2.1 Dataset Curation: SDXL and PixArt-α

To evaluate SDXL and PixArt-α, we first used a text generation large language model (*Anthropic, claude-3-opus-20240229*) to write 100 unique prompts for our image generation models. We standardized these prompts across models. Additionally, we used predefined random seeds across both models to ensure study repeatability, where a random seed corresponded to a repetition. All other parameters were kept consistent across runs: the width and height were set to 768 pixels, half of the maximum resolution available. The scheduler used was K-Euler. The guidance scale was set to 7.5 and the number of inference steps was set to 20. For SDXL, no refiner was used. Model inference was run on an Nvidia A40 hosted on Replicate.

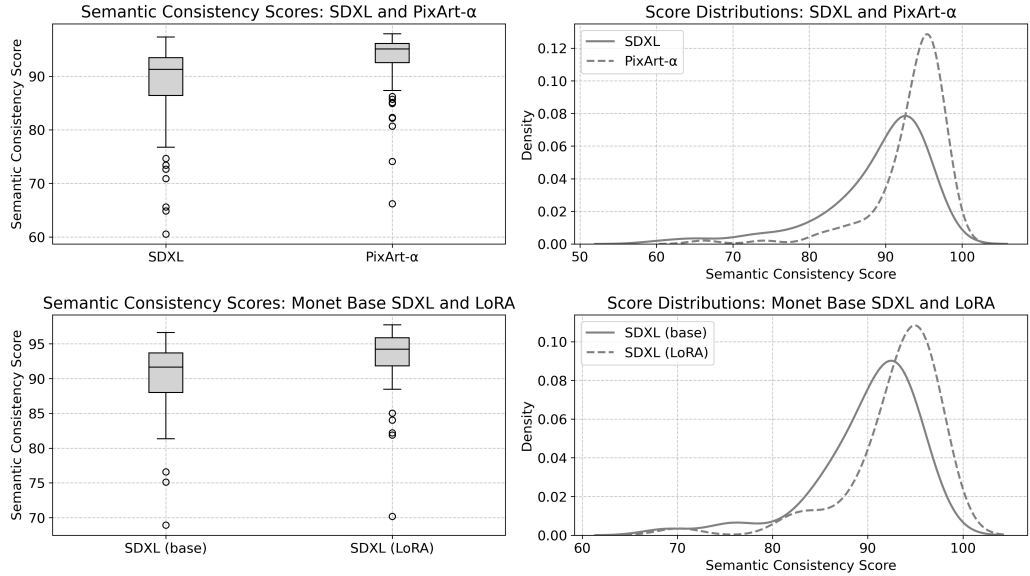

Figure 2: (top row) SDXL and PixArt-α show significant differences in their paired scores and distributions; visualized with boxplots and kernel density estimation plots. (bottom row) SDXL (base) and SDXL (LoRA fine-tuned on Monet) show significant differences in their paired scores and distributions; visualized with boxplots and kernel density estimation plots.

### 3.2.2 Dataset Curation: SDXL and LoRA

To investigate the impact of low-rank adaptation (LoRA) fine-tuning on SDXL, we fine-tuned the weights of SDXL using low-rank adaptation on nine Monet paintings from the public domain. Input images were processed using SwinIR (upscaling), BLIP (captioning), and CLIPSeg for removing regions of images not helpful for training (temperature 1.0). The batch size was 4, epochs were 1000, U-Net's learning rate was 1e-6, textual inversion's learning rate scaling was 3e-4, and LoRA embeddings' learning rate scaling was 1e-4. LoRA fine tuning was run on an Nvidia A40 hosted on Replicate.

We utilized a subset of 50 of the prompts used for the dataset curation of the SDXL and PixArt-α model comparison and used the same random seeds across models. Prompts were modified for input into SDXL with "in the style of Monet" appended to the end. Similarly, prompts were modified for the LoRA model to append "in the style of TOK", where "TOK" is the unique token string that was used during training to refer to the concept in the input images, the style of Monet's paintings.

### 3.2.3 Human Annotation

Annotation was performed by 13 human annotators. We built an annotation interface that displayed galleries of images generated by SDXL and PixArt-α side by side, and annotators selected the gallery they believed had the highest consistency and cycled through each prompt. Agreement was measured by comparing the model with the highest semantic consistency score to each annotator's choices and the collective response from all annotators (aggregated by frequency).

### 3.2.4 Sensitivity Analysis

We conducted a sensitivity analysis to determine the optimal number of prompt repetitions for our analysis to ensure a balance between accuracy and computational efficiency. We computed the pairwise mean CLIP score for 10 different prompts for both SDXL and PixArt-α. We computed this score for 10, 20, 30, 40, 50, 60, 70, 80, 90, and 100 repetitions (random seeds).

Based on our findings from the sensitivity analysis, which are detailed in the Experiments section, we performed 20 repetitions of each prompt for each model, resulting in 20*100*2 = 4k images for analysis.

### 3.2.5 Statistical Analysis

The pairwise mean CLIP score was calculated for each prompt and each model. A Kolmogorov-Smirnov test of normality was used to determine that the distributions of scores for each model were not normal ($p<0.05$). Accordingly, statistical significance was examined using a Wilcoxon signed-rank test (non-parametric paired sample test of significance) and a two-sample Kolmogorov-Smirnov test (non-parametric test used to determine if two samples are drawn from the same continuous distribution).

All code and data used in this paper have been open-sourced here: *https://github.com/brinnaebent/semantic-consistency-score.*

## 4. Experiments

### 4.1. Sensitivity Analysis

We conducted a sensitivity analysis to determine the optimal number of prompt repetitions for our analysis to ensure a balance between accuracy and computational efficiency. We found that a minimum of 20 repetitions was needed to ensure the score was within 1% of the mean score across all repetitions and within 1% of the score obtained with 100 repetitions. In 95% of the iterations, using 20 repetitions resulted in a score within 0.5% of both the mean score across all repetitions and the score obtained with 100 repetitions.

### 4.2. Model Comparison: SDXL and PixArt-α

We explored differences in image generation consistency between SDXL and PixArt-α, two state of the art open-source models (Figure 2). Across 100 prompts and 2k images per model, SDXL had a mean consistency score of 88.9±7.1 (median 91.3) and PixArt-α had a mean consistency score of 93.4±4.9 (median 95.1). The two-sample Kolmogorov-Smirnov test showed significant differences between the distributions of scores of the two models (KS statistic=0.48; p-value=8.44e-11). The Wilcoxon signed-rank test also showed significant differences between the paired scores (Wilcoxon statistic=110.0; p-value=1.01e-16).

The comparison between human annotations and the highest semantic consistency score revealed a high degree of agreement. The model with the highest semantic consistency score matched the most common selection among human annotators 94% of the time. Across all annotators, there was an average agreement rate of 90.9% [range 86%-94%].

### 4.3. Model Comparison: SDXL and fine-tuned SDXL with LoRA

We explored differences in image generation consistency between base SDXL and a LoRA fine-tuned version of SDXL. Across 50 prompts and 1k images per model, SDXL had a mean consistency score of 90.1±5.4 (median 91.7) and the LoRA fine-tuned SDXL model a mean consistency score of 92.9±5.0 (median 94.2). The two-sample Kolmogorov-Smirnov test showed significant differences between the distributions of scores of the two models (KS statistic=0.38; p-value=0.001). The Wilcoxon signed-rank test also showed significant differences between the paired scores (Wilcoxon statistic=95.0; p-value=5.80e-09).

### 4.4. Limitations

This study would be greatly augmented by further comparison to human judgment of consistency of image generation. Furthermore, we used the CLIP embedding model because it has been shown in other use cases to be robust [4, 8-9]; however, other multimodal embedding models, such as BLIP2 [17], should be explored in future work, especially given that CLIP models have been shown to pick up biases from input prompts [18].

## 5. Conclusions

The primary objective of this paper is to highlight the importance of measuring the consistency, or repeatability, of image generation models and to suggest a semantic method for doing so. In this paper, we propose using a Semantic Consistency Score based on a pairwise mean CLIP score. We then use this metric to compare two state of the art open-source models, SDXL and PixArt-α, in addition to SDXL and a LoRA fine-tuned SDXL.

This evaluation highlighted PixArt-α's superior consistency and reduced variability across prompts compared to SDXL. These findings have important implications for selecting the appropriate model for specific tasks: in applications demanding high consistency, opting for the more consistent PixArt-α model would be more advantageous. Conversely, when a diverse output is required, SDXL would be a more suitable choice. By precisely quantifying this consistency, we can better differentiate between models and make informed decisions regarding the selection of model architectures for different use cases.

LoRA fine-tuning of diffusion model weights is a popular approach to creating models that are more aligned to desired outputs. Through our exploration with our Semantic Consistency Score, we showed that our LoRA fine-tuned version of SDXL was more semantically consistent than base SDXL. The Semantic Consistency Score we have proposed in this study is a measure of desired image generation alignment and could be used in evaluation of LoRA models for specific tasks.

Other use cases include the evaluation of prompts, which could be useful when attempting to quantify and codify prompt engineering for various use cases, including cohesive story and movie generation using image generation. The idea of quantifying the consistency of generative model outputs could be extended beyond image generation to other modalities, such as evaluating consistency of generated text or audio-based outputs.